\documentclass[a4paper,conference]{IEEEtran}
\usepackage[switch]{lineno} 
\usepackage{graphicx}
\usepackage{enumitem}
\usepackage{changepage,threeparttable}
\usepackage{booktabs, multirow}
\usepackage{soul}
\usepackage[table]{xcolor} 
\usepackage{cite}
\usepackage{amsmath}
\usepackage{amssymb}
\usepackage[bookmarks=false]{hyperref}
\begin{document}
\title{Attention based Writer Independent Verification}

\author{
\IEEEauthorblockN{Mohammad Abuzar Shaikh\IEEEauthorrefmark{1}, Tiehang Duan\IEEEauthorrefmark{2}, Mihir Chauhan\IEEEauthorrefmark{3} and Sargur N. Srihari\IEEEauthorrefmark{4}}
\IEEEauthorblockA{Department of Computer Science and Engineering\\
State University of New York at Buffalo\\
Email: \IEEEauthorrefmark{1}mshaikh2@buffalo.edu, \IEEEauthorrefmark{2}tiehangd@buffalo.edu, \IEEEauthorrefmark{3}mihirhem@buffalo.edu, \IEEEauthorrefmark{4}srihari@buffalo.edu}
}

\maketitle

\begin{abstract}
The task of writer verification is to provide a likelihood score for whether the queried and known handwritten image samples belong to the same writer or not. Such a task calls for the neural network to make it's outcome interpretable, i.e. provide a view into the network's decision making process.  We implement and integrate cross-attention and soft-attention mechanisms to capture the highly correlated and salient points in feature space of 2D inputs. The attention maps serve as an  explanation premise for the network's output likelihood score. The attention mechanism also allows the network to focus more on relevant areas of the input, thus improving the classification performance. Our proposed approach achieves a precision of 86\% for detecting intra-writer cases in CEDAR cursive ``AND" dataset. Furthermore, we generate meaningful explanations for the provided decision by extracting attention maps from multiple levels of the network.
\end{abstract}


%
\IEEEpeerreviewmaketitle

\section{Introduction}
Writer independent verification is the task to measure the similarity of two given handwritten samples as how likely is it that the samples were written by the same, without having any knowledge of writers identity. There have been many efforts in this field to provide an automated hint to streamline the job of manual handwriting examiners, making this an interesting research problem. 
\newline \indent A general intuition is that samples from a single source tend to be similar while samples from different sources tend to show bigger variances. The premise for finding unique characteristics is based on the hypothesis that every individual has a unique way of writing \cite{srihari_individuality_2001}. Further studies show that two different writers may also happen to have a similar writing style. This makes the problem of handwriting verification challenging.
\newline \indent With the advent of automated pattern learning methods; especially with models that can be trained to focus on key areas, it is possible to design a robust system to assist a Forensic Document Examiner (FDE). In this work we propose attention based approaches for the task of handwriting verification. We produce two kind of attention maps\footnote{Code is publicly available on: \\ \href{https://github.com/mshaikh2/AttentionHandwritingVerification}{https://github.com/mshaikh2/AttentionHandwritingVerification}} which highlights (i)  the important corresponding pixel regions between two images that the network deemed similar (ii) the high correlated pixel regions, in the feature space of given samples, that the network attends to provide its decision. Such visualizations render the desired interpretability for forensic verification systems.
\section{Related Work}
The task of verification is prevalent across bio-metric domains, viz. face verification in DeepFace \cite{taigman_deepface_2014}, fingerprint verification \cite{jain_-line_1997}, handwritten evidence verification \cite{gideon_handwritten_2018}, iris verification \cite{nguyen_iris_2018}, speaker verification \cite{speaker_verification}. Our paper focuses on handwritten evidence verification and proposes a novel approach for writer independent verification.
\newline \indent Earlier, researchers in the field of handwriting verification used handcrafted features coupled with classic learning techniques, where, Réjean et al \cite{PLAMONDON1989107} present an overview of pre-processing techniques and feature extraction methods from handwritten text, Srihari et al introduced Gradient Structural Concavity (GSC) feature extraction technique \cite{srihari_individuality_2001} for writer identification and verification and present CEDAR-FOX tool, which is a noteworthy contribution in this field. L Bovino et al \cite{bovino2003multi} present a multi-expert system based on stroke-oriented description for dynamic verification, Marius Bulacu et al \cite{bulacu_text-independent_2007} present statistical methods that operate on the texture level and the character-shape (allograph) level, AA Brink \cite{brink_towards_2011} prove that slantness as a feature for handwritten text is overrated which shows the robustness of handcrafted features, K Tselios et al \cite{tselios2011automated} present automated feature extraction method  based for handwritten text, D. Bertolini et al\cite{BERTOLINI20132069} discuss the use of texture descriptors to perform writer verification, M. N. Abdi et al\cite{abdi_model-based_2015} propose a grapheme-based approach to offline Arabic writer identification and verification, Manabu Okawa et al\cite{okawa_text_2015} propose a text and user generic model for writer verification that uses a combination of pen pressure information from ink intensity and writing indentations. 
\newline \indent However, the recently proposed deep learning models enabled automatic extraction of generic features. Ameur Bensefia\cite{bensefia_writer_2016} use Levenshtein edit distance based on Fisher-Wagner algorithm to estimate the cost of transforming one handwritten word into another, Shaikh et al\cite{shaikh_hybrid_2018} present a Hybrid Deep Learning architecture combining handcrafted features and Convolutional Neural Network (CNN) based features, Chu et al\cite{chu_writer_2018} propose an end-to-end deep learning method based on statistical features extracted on set-of-samples level.  
\newline \indent Nevertheless, verification is still a challenge, as lack of interpretability still exists in these current widely used models. This makes it tough for the FDE to rely on model's binary decision. There is a demand to provide visual or textual explanations for a models decision which can aid the FDE's in verification of the samples confidently. Chauhan et al\cite{chauhan_explanation_2019} generate explanations for the confidence provided by CNN by mapping the input image to 15 annotated features provided by experts. However, the generation of such annotated data using using manual labor takes significant effort and is time consuming. 
\newline \indent Hence, there is a need for a model to display what location is it attending to, while making a decision. Attention models proposed by Dzmitry Bahdanau et al\cite{BahdanauCB14} automatically search for parts of a source input that are relevant to predicting the output, Luong et al\cite{luong_effective_2015} proposed hard and soft attention wherein hard attention allows the output to be influenced by exactly one most relevant input, and soft attention uses a less strict criteria by only boosting the most relevant input while still allowing a subset of other inputs to contribute towards the models decision. Vaswani et al\cite{vaswani_attention_2017} presented a very powerful model in Natural Language Processing (NLP) based solely on multi headed attention mechanisms, dispensing with recurrence and convolutions entirely. Han Zhang et al\cite{zhang_self-attention_2019} extend the concept of attention to 2d images to capture non-local dependencies and demonstrate the models efficacy generating images. Furthermore, Naofumi Tomita et al\cite{tomita_attention-based_2019} leverage 3D CNN to perform object localization using attention. 
\newline \indent In this work, we propose to tackle the writer verification task with Multi-Head Cross Attention (MHCA) mechanism, where the model compares the two input images and focuses on the corresponding and relevant pixel in their feature space. The network integrates a Soft Attention (SA) mechanism designed using 3D CNN to help it attend more on the important correlated features for classification. We then generate two attention maps, (i) to show which locations in the two images are highly correlated, (ii) to show at which locations the model focuses for classification. We conduct extensive experiments on two datasets to show the effectiveness of cross attention combined with soft attention. Our experiments demonstrate that the approach performs at par or better than the existing state of the art methods on widely used datasets while also providing insightful explanation on model's decision.
\section{Dataset}
We perform experiments on ``AND" Dataset \cite{srihari_individuality_2001, chauhan_explanation_2019} and test the performance of best model on CEDAR Signature \cite{signature_verif_dataset} dataset. Samples in both the datasets were handwritten on a paper and then scanned to create images of the manuscripts. For both datasets, after appropriate square padding corresponding to the maximum width and height of the sample, we resize each of them to have consistent size of $64 \times 64$  using bi-cubic interpolation. Moreover, we invert the pixels such that all the background pixels are 0. We apply a threshold on the images in CEDAR Signature dataset and change any pixel values which less than 30 to 0. Finally, we normalize the images in both datasets by dividing each pixel with $255$ so that the range of values of the pixels stay between $0$ and $1$.
\subsection{Cursive ``AND''}
The dataset is formed by cropping segments containing the word ``AND" from the full letter CEDAR dataset \cite{srihari_individuality_2001}. Post cropping, some non-``AND" words were removed manually. After cleaning, we have a dataset of 1533 writers, constituting around 14000 samples variable shapes.
\newline \indent We define the samples written by the same and different writer as ``intra-writer" and ``inter-writer'' pairs respectively.
\begin{figure}[!h]
\begin{center}
\includegraphics[width=0.48\textwidth]{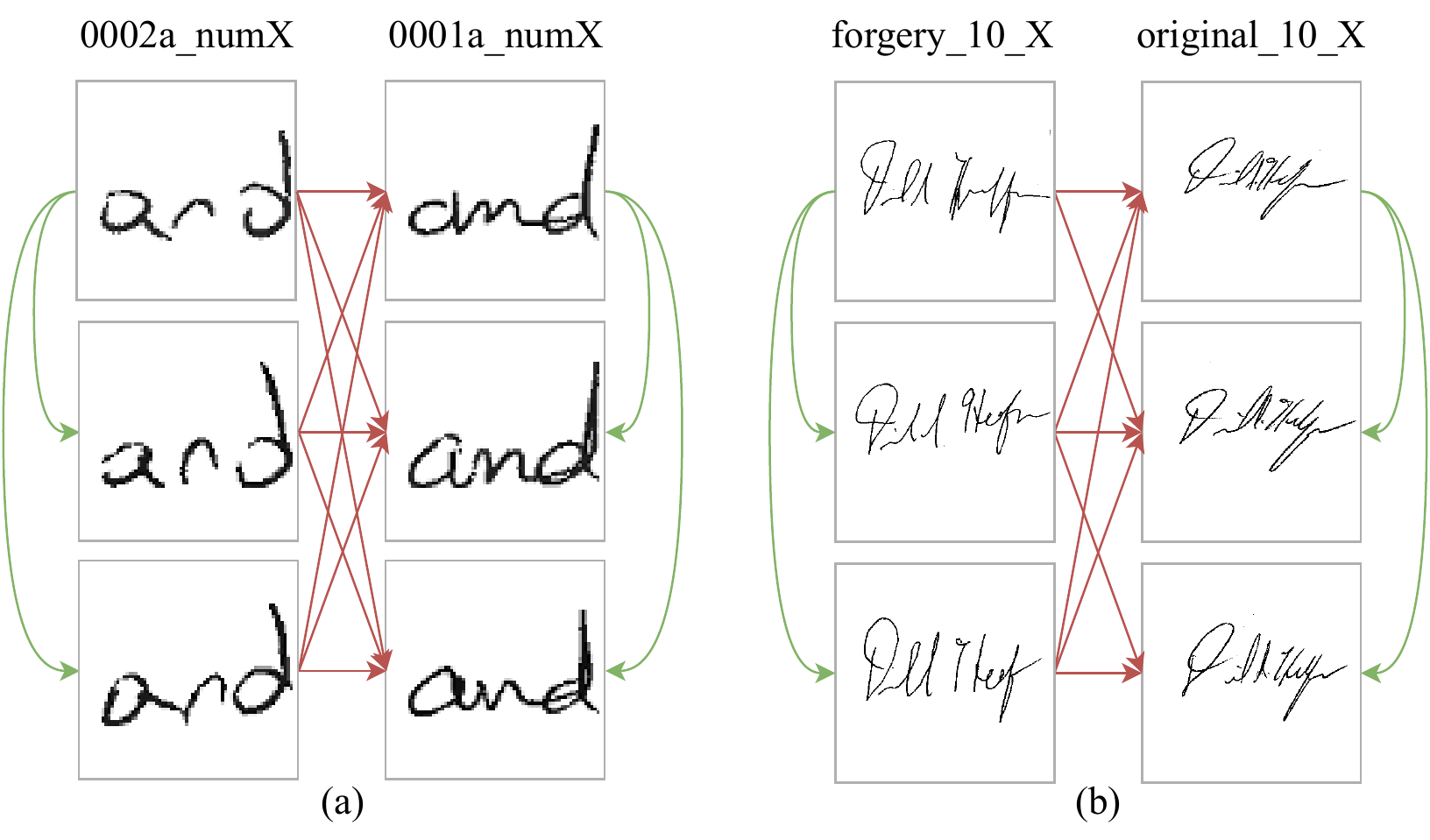}
\caption{\label{fig:dataset} (a) Samples from the CEDAR cursive ``AND'' dataset. (b) Samples from CEDAR Signatures dataset. The green arrows indicate pairing of samples from the same writer. The red arrows indicate pairing of samples from different writers. The character X $ \in \{1,2,3\} $} 
\end{center}
\end{figure}
\newline \indent Each writer on an average has 9 samples of the word ``AND''. Hence we have around $ {{9 \choose 2} = 36}$ samples of similar pair per writer. Therefore, we have around ${1533 \times 36 = 55188}$ pairs of intra-writer samples. Furthermore, we shuffle all the samples and use K-Fold cross validation \cite{k_fold_95} with ${K=5}$. We choose 4 folds as training set and 1 fold as testing set. Within the respective folds, we randomly generate 10 times more inter-writer pairs than intra-writer pairs, to make the training more effective as done in  \cite{kurczab_influence_2014}. Thus, the ratio of intra-writer pairs to inter-writer pairs in our training and testing sets is 1:10 for all experiments with this dataset.
\subsection{CEDAR signature}
CEDAR signature database \cite{signature_verif_dataset} contains signatures from 55 individuals. Each  of  these  signers  signed  24  genuine  signatures.  For each signer there are 24 forgery samples from about 20 skillful forgers.   Hence the  dataset contains 1,320 genuine signatures as well as 1,320 forged signatures. All images in this dataset are available in gray scale mode.
\newline \indent We divide this dataset in 11 folds based on writer ids. Next, we consider random 10 parts as training set and remaining 1 part as test set. Thus, there are $50$ writers are in training set and $5$ writers in testing set. Furthermore, we have ${24 \choose 2}= 276$ genuine pairs per writer with $50 \times 276 = 13800$ and $5\times276=1380 $ genuine-genuine pairs in training and in testing set respectively. Furthermore, we consider all the $24 \times 24 \times 50 = 28800$ and $24 \times 24 \times 5 = 2880$ genuine-forgery pairs for training and testing set respectively. Hence, the ratio of genuine-genuine-writer pairs to forgery-genuine-writer pairs in our training and testing sets is almost 1:2 for this dataset.

\section{Method}
We employ cross attention (CA) and soft attention (SA) mechanism in conjunction with Inception-Resnet-v2 (IRv2) \cite{szegedy_inception-v4_2017}, a powerful feature extractor. The overall network is displayed in Fig.  \ref{fig:architecture}. We first process the input through the original stem block of IRv2 or two convolutional blocks of VGG16 (we consider this as VGG16's stem). The stem block extracts $d$ features while scaling down the input height $H$ and width $W$ to $h'$ and $w'$ respectively. The higher level modules are discussed below:
\subsection{Shared Weights}
To process two input images simultaneously we train two copies of the stem block in a Siamese setting, such that both the branches have shared weights. The stem block outputs a 3D tensor $\mathbb{R}^{h'\times w'\times d}$. The stem is the replica of the IRv2 stem block, with only difference that inputs are gray scaled and of size $64 \times 64$.
\subsection{Cross Attention}

To cross attend $f_1$, $f_2$ the feature outputs of stem 1 and stem 2 respectively, we first generate the key $k$, value $v$ and query $q$.
\begin{figure}[!htb]
\begin{center}
\includegraphics[width=0.5\textwidth]{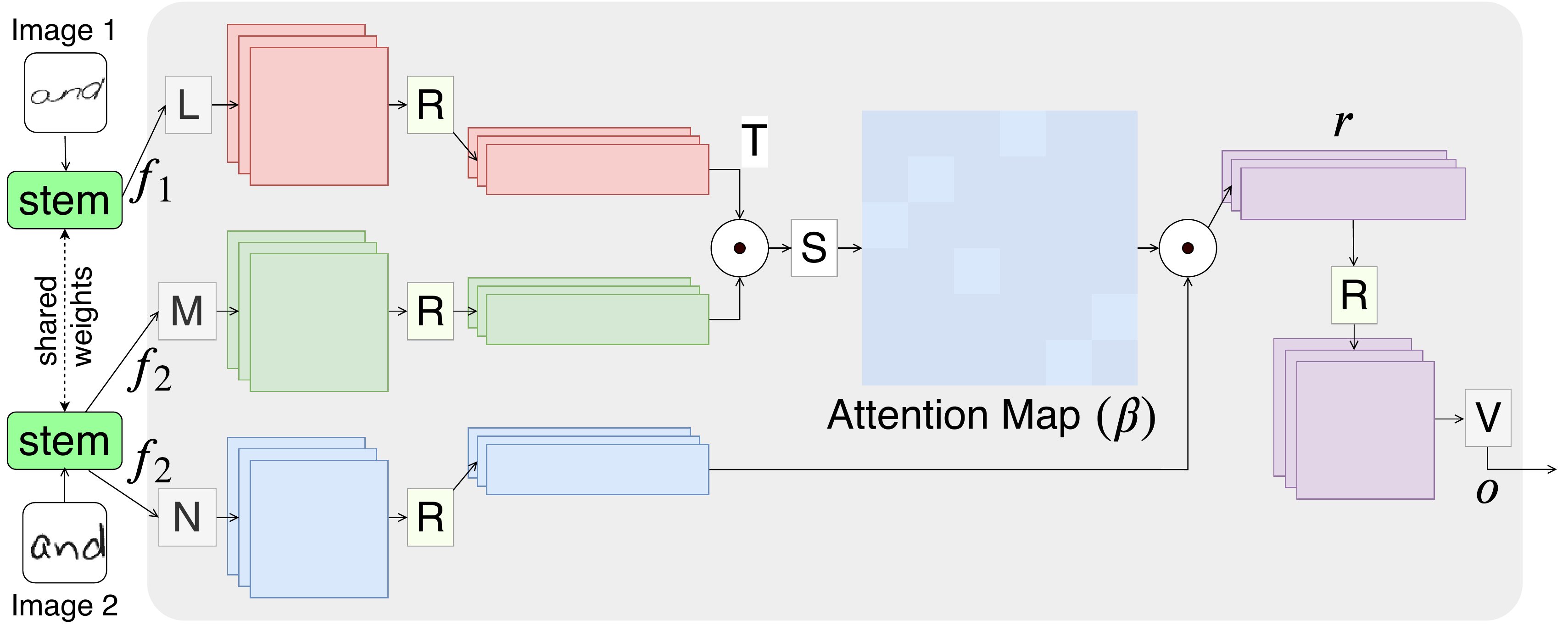}
\caption{\label{fig:cross_attention} Internal working of a Cross Attention module. L, M, N and V are separate 1 $\times$ 1 convolutions. R is the reshape operation. $\odot$ represents matrix multiplication. S denotes softmax which is performed per row to generate Attention Map.} 
\end{center}
\end{figure}Where $k = M(f_2)$; $v = N(f_2)$; $q = L(f_1)$ and $M,N,L$ are $1 \times 1$ convolution functions. Evidently in Fig. \ref{fig:cross_attention}, image 2 is considered as key input and image 1 as the query input.
The $k$, $q$ and $v$ are then reshaped to tensor $\in \mathbb{R}^{h' * w' \times d} = \mathbb{R}^{t \times d}$. Cross Attention (CA) is inspired by the self-attention presented in \cite{zhang_self-attention_2019} and is demonstrated in Fig. \ref{fig:cross_attention}. We calculate the relevance between $q$ and $k$, which are features of two distinct inputs. 

\begin{equation}
    \beta_{i,j} = \frac{exp(z_{ij})}{\sum_{j=1}^{t} exp(z_{ij})}, \text{where } z_{ij} = k \odot q^{T}; z_{ij} \in \mathbb{R}^{t \times t}
    \label{eq:cross_attention}
\end{equation}
Here, $\odot$ represents matrix multiplication; $z_{ij}$ calculates the relevance between every $i^{th}$ and $j^{th}$ location in k and q respectively. $\beta_{ij}$, the attention map represents softmax normalization across every $i^{th}$ row as Eq. \ref{eq:cross_attention} boosts the most relevant $j^{th}$ value and suppresses the non-relevant values, for  each corresponding $i^{th}$ row .
Next, we calculate the enhanced representation $r$, of image 1 with infused contextual representation of image 2.
\begin{equation}
    r_{i} = \sum_{j=1}^{t} \beta_{i,j}v_{j}, \text{where } r_{i} \in \{r_{1}, r_{2}, \dots, r_{t}\}
    \label{eq:context_vector}
\end{equation}
$r$ is then reshaped to tensor $\in \mathbb{R}^{h' \times w' \times d}$ and the final output vector $o$ is calculated. Here, $o = V(r)$, where $V$ is $1 \times 1$ convolution function.
\newline \indent Furthermore, we swap the inputs and perform the aforementioned calculations again. This recalculation makes sure that features of image 1 have the context of image 2 and vice-versa. Finally, the two outputs are concatenated on channel axis such that the resultant output $\in \mathbb{R}^{h' \times w' \times 2d}$
\newline \indent The above operations constitute two attention heads, that is one set of weights for each output. We can scale this module to have $n$ attention heads. In our experiments $n=8$, such that first four heads are setup with image 1 as key and remaining four have image 2 as key. Multiple attention heads help the network to identify more than one relevant feature locations of one image with respect to a given index in another.
\subsection{Soft Attention}
The foreground pixels that contain a handwritten stroke are useful pixels. Each sample in the data contains only $7\%$ percent of such useful pixels, as rest of the pixels are background with no information. Inspired by the work done by \cite{tomita_attention-based_2019} we propose a soft attention technique that uses 3D-convolution \cite{tran_learning_2015} to attend and identify only the most important features responsible for classification. 
\begin{figure}[!htp]
\begin{center}
\includegraphics[width=0.5\textwidth]{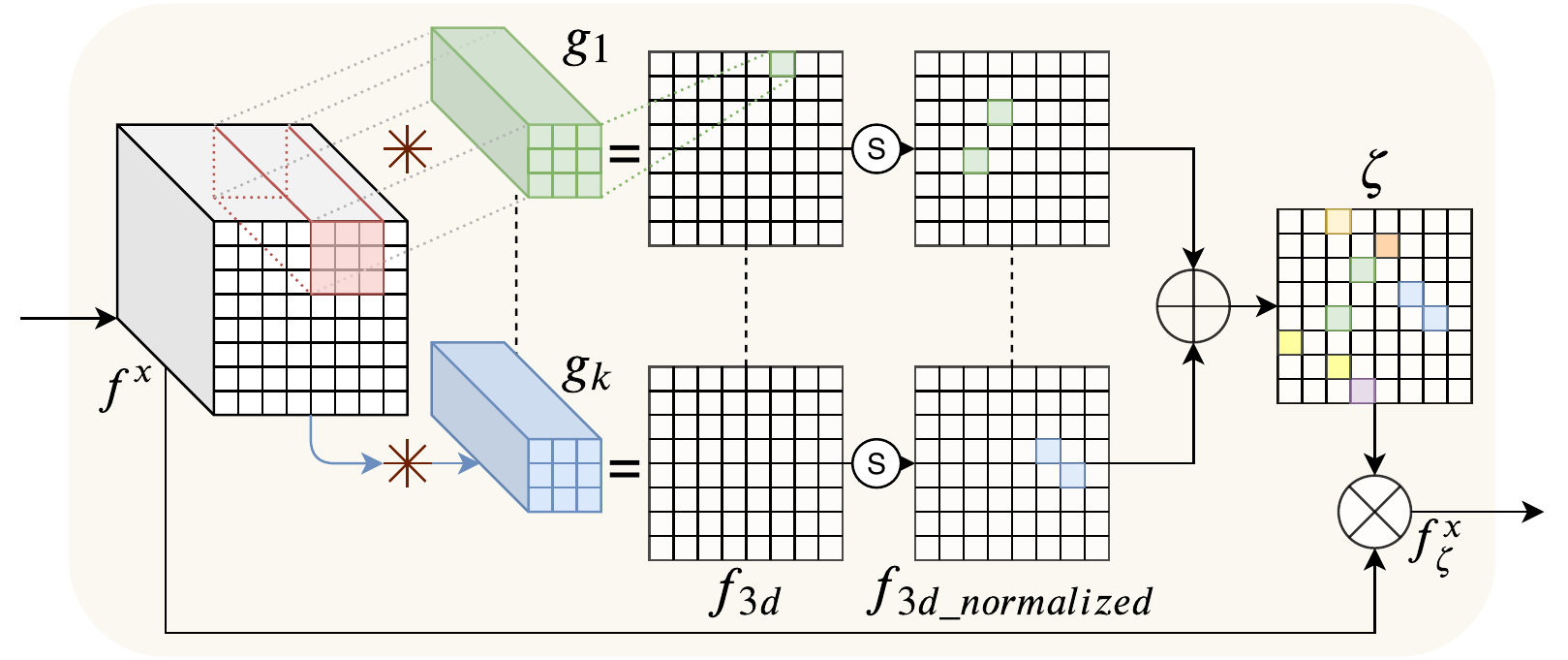}
\caption{\label{fig:soft_attention} Internal working of a Soft Attention module. $\oplus$ denotes aggregation and $\ast$ represents convolution operation. $S$ denotes softmax which is performed over one feature map to generate one attention map.} 
\end{center}
\end{figure}
\newline \indent Convolving a given three dimension output $f^x \in \mathbb{R}^{h^x \times w^x \times d^x}$ with one 3D kernel $g$ of size $3 \times 3 \times d^x$, generates a feature map $f_{3d} \in \mathbb{R}^{h^x \times w^x \times 1}$.
\newline \indent Having $K$ such kernels represents $K$ attention heads and generates a feature map $f_{3d} \in \mathbb{R}^{h^x \times w^x \times K}$. Next, we normalize each of the $K$ feature maps and aggregate them to calculate the soft attention scores $\zeta$:
\begin{equation}
    \zeta = \sum_{k=1}^{K}{\frac{exp(f_{3d_{ij}})}{\sum_{i=1}^{w^x}\sum_{j=1}^{h^x}{exp(f_{3d_{ij}})}}}; \text{ where } f_{3d} = g(f^x)
    \label{eq:3d_attention}
\end{equation}
The normalization assigns an importance score for relevant locations in the feature map. Next we multiply $f^x$ with $\zeta$ and obtain $f_{\zeta}^{x}$, to scale the values of salient locations. Finally, $o_{\zeta}$ is calculated by adding $f^x$ with the product of $f_{\zeta}^{x}$ and $\omega$, where $\omega$ is a learnable scalar.
\begin{equation}
    o_{\zeta} = f^x + \omega f_{\zeta}^{x}
    \label{eq:3d_attention_output}
\end{equation}
This enables the network to decide how much attention should be applied over specific locations of the feature map.
\begin{figure}[!htp]
\begin{center}
\includegraphics[width=0.3\textwidth]{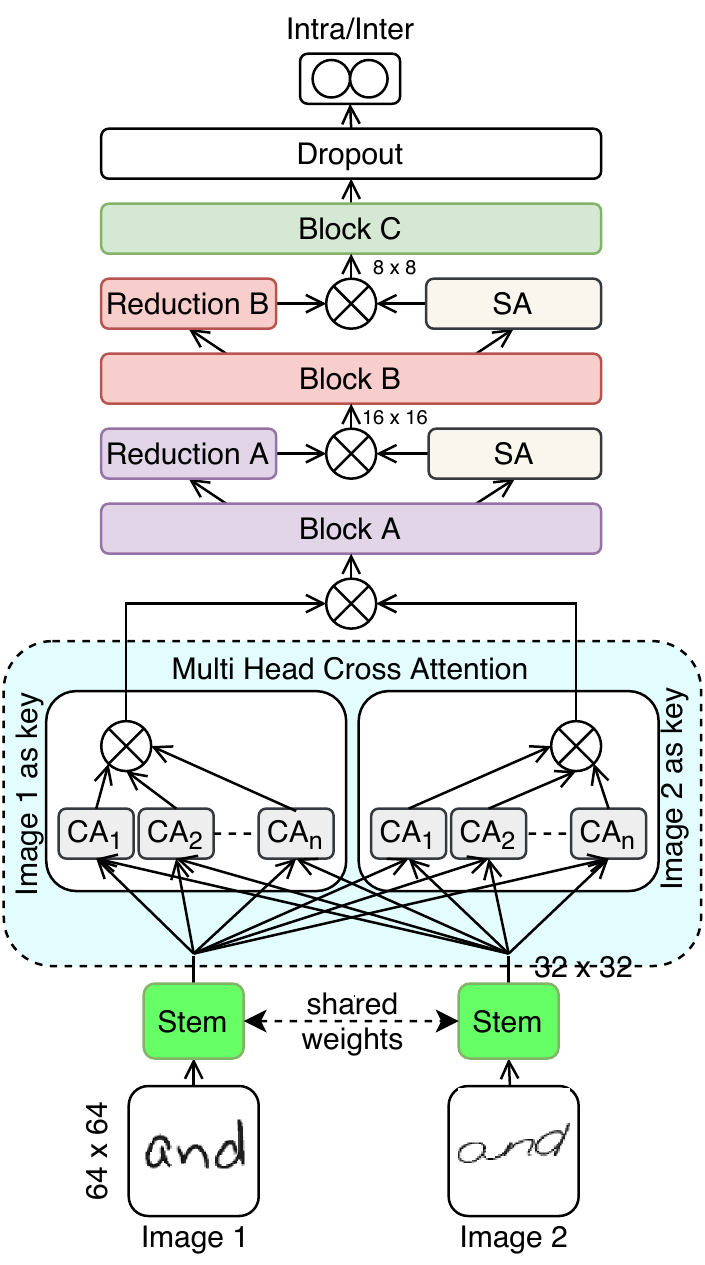}
\caption{\label{fig:architecture} End to end architecture using VGG16\cite{vgg16} or Inception-Resnet-v2 (IRv2) \cite{szegedy_inception-v4_2017} blocks, multiple Cross Attention (CA) and Soft Attention (SA) modules. $\otimes$ indicates concatenation on channel axis.}
\end{center}
\end{figure}
\subsection{Loss Function}
\begin{table*}[!htp]\centering
\caption{Establishing baseline on ``AND" dataset and parameter selection for FL}\label{tab:baseline_and}
\scriptsize
\begin{tabular}{lc|cccccc|ccccccc}\toprule
& &\multicolumn{6}{c|}{Concat Setting} &\multicolumn{6}{c}{Siamese Setting} \\\cmidrule{3-14}
\textbf{$\alpha$} &\textbf{$\gamma$} &\textbf{F-1} &\textbf{P} &\textbf{R} &\textbf{FAR} &\textbf{FRR} &\textbf{Acc} &\textbf{F-1} &\textbf{P} &\textbf{R} &\textbf{FAR} &\textbf{FRR} &\textbf{Acc } \\\midrule
\multirow{7}{*}{0.50} &0.00 &0.72 &0.85 &0.62 &1.12 &37.53 &95.56 &0.78 &0.83 &0.74 &1.69 &25.99 &95.74 \\
&0.10 &0.71 &0.85 &0.61 &1.05 &39.11 &95.48 &0.77 &0.91 &0.66 &0.94 &33.63 &96.82 \\
&0.20 &0.73 &0.83 &0.65 &1.38 &34.79 &95.58 &0.77 &0.88 &0.69 &1.07 &31.48 &96.92 \\
&0.50 &0.74 &0.83 &0.67 &1.35 &33.09 &95.76 &0.79 &0.89 &0.71 &1.10 &29.06 &97.10 \\
&1.00 &0.73 &0.88 &0.62 &0.83 &37.78 &95.80 &0.81 &0.93 &0.72 &1.71 &27.99 &97.14 \\
&2.00 &0.74 &0.82 &0.67 &1.47 &33.31 &95.63 &0.79 &0.88 &0.73 &1.45 &27.38 &96.97 \\
&5.00 &0.73 &0.77 &0.68 &2.00 &31.69 &95.29 &0.75 &0.83 &0.69 &1.02 &31.19 &96.63 \\
\midrule
\multirow{6}{*}{\textbf{0.75}} &0.10 &0.71 &0.83 &0.62 &1.33 &37.56 &95.37 &0.77 &0.88 &0.68 &0.67 &32.54 &96.71 \\
&0.20 &0.73 &0.85 &0.64 &1.13 &36.09 &95.69 &0.79 &0.91 &0.69 &0.92 &31.03 &97.03 \\
&0.50 &0.74 &0.83 &0.67 &1.42 &32.66 &95.73 &0.79 &0.88 &0.72 &1.07 &28.01 &97.07 \\
&1.00 &0.72 &0.83 &0.64 &1.31 &35.69 &95.56 &0.79 &0.89 &0.71 &1.70 &28.91 &96.89 \\
&\textbf{2.00} &\textbf{0.74} &\textbf{0.81} &\textbf{0.68} &\textbf{1.59} &\textbf{32.41} &\textbf{95.60} &\textbf{0.81} &\textbf{0.87} &\textbf{0.76} &\textbf{1.77} &\textbf{24.33} &\textbf{96.94} \\
&5.00 &0.72 &0.86 &0.62 &1.00 &38.46 &95.58 &0.79 &0.92 &0.69 &1.36 &30.86 &96.92 \\
\bottomrule
\end{tabular}
\end{table*}
This problem is set up as a two class classification problem. Hence, we optimize the network using categorical cross-entropy (CCE) function 
\begin{equation}
  \label{eq:multipleentropy}
    L_{CCE} =  - \log(p_t)
\end{equation}

However, since there is a large skew between inter and intra-class samples, we also experiment with Focal Loss (FL) \cite{lin_focal_2020} and implement the categorical focal loss $L_{fl}$ as
\begin{equation}
  \label{eq:focalloss}
    L_{FL} =  - \alpha_t (1-p_t)^{\gamma}log(p_t); 
\end{equation}
where \begin{equation*}
    p_t = 
    \begin{cases} 
        p & \text{if } y = 1 \\
        1-p & \text{otherwise}
    \end{cases};
     \alpha_t = 
    \begin{cases} 
        \alpha & \text{if } y = 1 \\
        1-\alpha & \text{otherwise}
    \end{cases}
\end{equation*}
 $p \in [0,1] $ is the model's estimated probability, and $y \in [0,1]$ is the ground truth label.
 
\section{Experiments and Evaluation}
\subsection{Setup}
The initial building blocks of our feature extraction model uses the feature extraction blocks from VGG16 \cite{vgg16} or Inveption-ResNetv2 (IRv2) model \cite{szegedy_inception-v4_2017}. These networks are the state-of-the-art image feature extractors trained on ImageNet \cite{imagenet} dataset for classification. 

We train the model in two settings of input: (1) Two images concatenated on channel axis (Concat), (2) Two images in parallel(Siamese).
In Concat setting, there is only one input to the network as we pre-process the two input images to overlay on one another. Soft-Attention mechanism is applicable to both the settings, and Cross-Attention is applicable only in case 2, when the two inputs are in a Siamese setting. Siamese setting is as displayed in Fig. \ref{fig:architecture}. 

The CA blocks are laid in parallel to extract key point correspondence between the input images. Next, a block, of SA layer followed by a $2 \times 2$ max-pool layer, is used alongside a standard feature reduction $2 \times 2$ max-pool block.
The outputs of the SA\_max-pool and standard max-pool are then concatenated on channel axis.
This helps the network attend more on important features for classification and reduce adherence to noisy part of data. 
To introduce some data augmentation and regularization we introduce a dropout of $0.5\%$ probability after each convolutional block.

\indent We train the network for 100 epochs with an early stopping patience of 20 epochs. Validation loss is monitored at every epoch to checkpoint the weights of the best model. For all the experiments we use Adam optimizer with learning rate, $lr = 0.0001$ and a decay, $lr_{decay}=1.0^{-6}$. Both the customized VGG16 and IRv2 networks contain around 23 million parameters respectively. We utilized three 11 GB Nvidia 1080Ti GPUs in parallel for all the experiments which were written in Keras framework on Tensorflow backend.

\subsection{Evaluation metrics}
We evaluate our models using F-1, Precision (P), Recall (R), False Acceptance Rate (FAR), False Rejection Rate (FRR) and Accuracy (Acc). FRR, also known as ``Type I" error, is the measure of the likelihood that the model will incorrectly classify an intra-writer sample as inter-writer. 
FAR, also known as ``Type II" error, is the measure of the likelihood that the model will predict an inter-writer sample as intra-writer. 
\subsection{Results}
Table \ref{tab:baseline_and} shows the performance of baseline models for ``Concat" and ``Siamese" setting without any attention modules. In Table \ref{tab:baseline_and} when $\gamma$ is set to $0$, FL is same as CCE.
We set $\alpha=0.75$ and $\gamma = 2.0$ and adopt Siamese setting for our further experiments.
\newline \indent Next, we experiment on the proposed attention based models and show the results in Table \ref{tab:results}. We also managed to replicate the work done in \cite{chauhan_explanation_2019} and report their metrics. As for work done by \cite{shaikh_hybrid_2018} we could only report their accuracy on this dataset. 
\begin{table}[!htp]\centering
\caption{Experiment Results of Various Models on ``AND" Dataset}\label{tab:results}
\scriptsize
\begin{tabular}{lccccccc}\toprule
\textbf{Method} &\textbf{F-1} &\textbf{P} &\textbf{R} &\textbf{FAR} &\textbf{FRR} &\textbf{Acc} \\\midrule
HDL \cite{shaikh_hybrid_2018} &- &- &- &- &- &92.16 \\
DAAM\_SAE \cite{chauhan_explanation_2019} &0.70 &0.85 &0.59 &3.69 &40.57 &95.23 \\
Concat\_Baseline &0.72 &0.85 &0.62 &1.59 &37.53 &95.60 \\
Concat\_SA &0.73 &0.85 &0.64 &1.73 &36.36 &95.69 \\
Siamese\_Baseline &0.78 &0.83 &0.74 &1.77 &26.00 &96.18 \\
Siamese\_CA\_SA &0.79 &0.84 &0.74 &1.93 &25.82 &96.34 \\
\textbf{Siamese\_MHCA\_SA} &\textbf{0.81} &\textbf{0.86} &\textbf{0.76} &\textbf{1.73} &\textbf{24.03} &\textbf{96.39} \\
\bottomrule
\end{tabular}
\end{table}
The MHCA combined with SA achieves the best results in Siamese setting. Moreover, the network is able to lower the FRR considerably as compared to vanilla settings. The difference in FAR and FRR values are due to the low number and high imbalance of samples per writer. We also experiment the application of MHCA on various levels of feature maps and display it's optimum location in Table \ref{tab:ca_postion}.
\begin{table}[!htp]\centering
\caption{Experiments to locate the optimum feature map size to add CA module}\label{tab:ca_postion}
\scriptsize
\begin{tabular}{lccccccc}\toprule
\textbf{FeatureMap Size} &\textbf{F-1} &\textbf{P} &\textbf{R} &\textbf{FAR} &\textbf{FRR} &\textbf{Acc} \\\midrule
16 $\times$ 16 &0.70 &0.78 &0.64 &1.84 &35.81 &95.06 \\
\textbf{32 $\times$ 32} &\textbf{0.81} &\textbf{0.86} &\textbf{0.76} &\textbf{1.73} &\textbf{24.03} &\textbf{96.39} \\
\bottomrule
\end{tabular}
\end{table}
We observe that as we increase the feature map size the performance improves. This is because as size reduces the features in the map become more abstract and there are less evidences to relate features of two images. \cite{zhang_self-attention_2019}
\begin{table}[!htp]\centering
\caption{Experiment Results of various Methods on CEDAR Signature Dataset}\label{tab:sign_compare}
\scriptsize
\begin{tabular}{lcccc}\toprule
\textbf{Method} &\textbf{FAR} &\textbf{FRR} &\textbf{Acc} \\\midrule
Graph matching \cite{chen_graph_signature} &8.20 &7.70 &92.10 \\
SigNet \cite{dey2017signet} &0.00 &0.00 &100.00 \\
SigNet-F (SVM) \cite{Hafemann_2017} &4.63 &4.63 &- \\
\textbf{Siamese\_MHCA\_SA} &\textbf{5.70} &\textbf{6.30} &\textbf{92.37} \\
\bottomrule
\end{tabular}
\end{table}
\newline \indent Furthermore, we train our model on CEDAR Signature \cite{signature_verif_dataset} to demonstrate that the proposed MHCA-SA mechanism is effective across different datasets. As displayed in Table \ref{tab:sign_compare} our method achieves comparable potential with most widely used methods. 
\subsection{Discussion and Ablation Analysis}
Writer verification data is inherently biased with significantly more negative samples overwhelming positive samples. We mainly rely on Precision, Recall and F-1 score due to their robustness to such imbalanced data. Accuracy and FAR on the other hand are sensitive to the ratio of negative samples. This may be the case in our experiments where an increase in the number of easy negative samples in test set can lead to a drastic drop in FAR and increase accuracy. 
\begin{figure}[!htp]
\begin{center}
\includegraphics[width=0.49\textwidth]{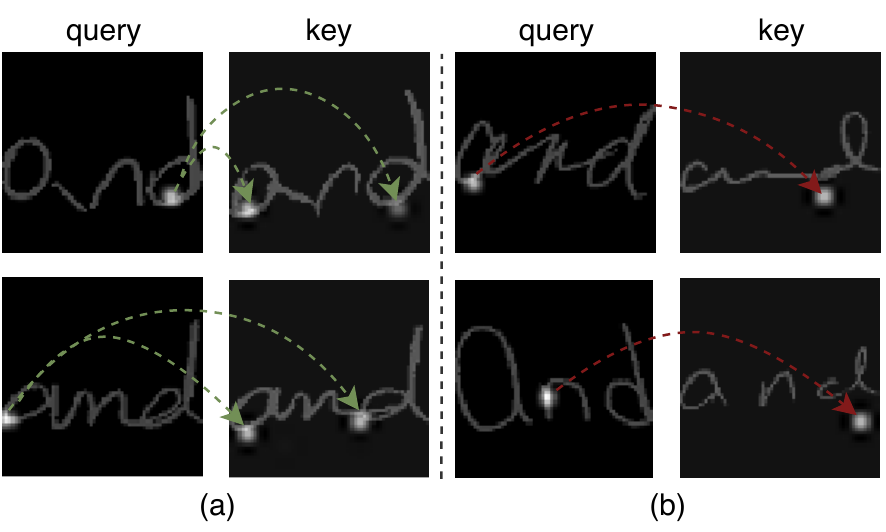}
\caption{\label{fig:ca_attentionmaps} CA Maps for input images. (a) CA Maps when inputs are are from the same writer, (b) CA Maps when inputs are from different writers.}
\end{center}
\end{figure}

Based on our observations, using VGG16 blocks performed better than employing IRv2 blocks, and hence, we report the results with VGG16 blocks.
Furthermore, we use \textit{softmax} in our classification layer rather than \textit{sigmoid} as softmax coupled with CCE tend to distribute the probability distribution far apart without requiring any constant margin.  Moreover, CA finds the corresponding related pixels between two given images using cosine similarity score in feature space which leads to further improvement in feature generation. Additionally, SA boosts the network's ability of important feature selection and hence performance.
\newline \indent We extract the attention maps from the CA and SA modules which provides interpretation on the models learning process and finding evidence of writer similarities. Fig. \ref{fig:ca_attentionmaps} displays attention energy maps from one of the CA heads. The images in Fig. \ref{fig:ca_attentionmaps}a and Fig. \ref{fig:ca_attentionmaps}b show the cross attention correspondence maps, when the images were from the same/different writer respectively. In both the subsections the images on the left are considered as the query and the adjacently right images are considered as key images to attend on. As shown in \ref{fig:ca_attentionmaps}a, when a pixel on the curve of ``a" is used as a query, two corresponding points from the key image are returned by the model. Visibly, these two points extracted from the key image are similar in shape and stroke. We hypothesize that the model uses this information to further strengthen it's belief that the images are from same writer. 

Furthermore, as seen in Fig. \ref{fig:ca_attentionmaps}b the model could not identify similar strokes, edges or contours, from the key image for the given query points, in the two images towards bottom right. This perhaps enhances the ability of the model to identify dissimilar writings. 

We observe the pattern, of matching corresponding key points in similar alphabets, uniform across intra-writer samples. Also, the phenomena of query pixel matching an exact same black/non-important pixel in the image, is consistent for inter-writer samples. This provides a hint to the FDE's for the reason behind the models decision.

Next, we extract the attention maps from the SA layers, to display the areas which the network has jointly attended.
\begin{figure}[!h]
\begin{center}
\includegraphics[width=0.49\textwidth]{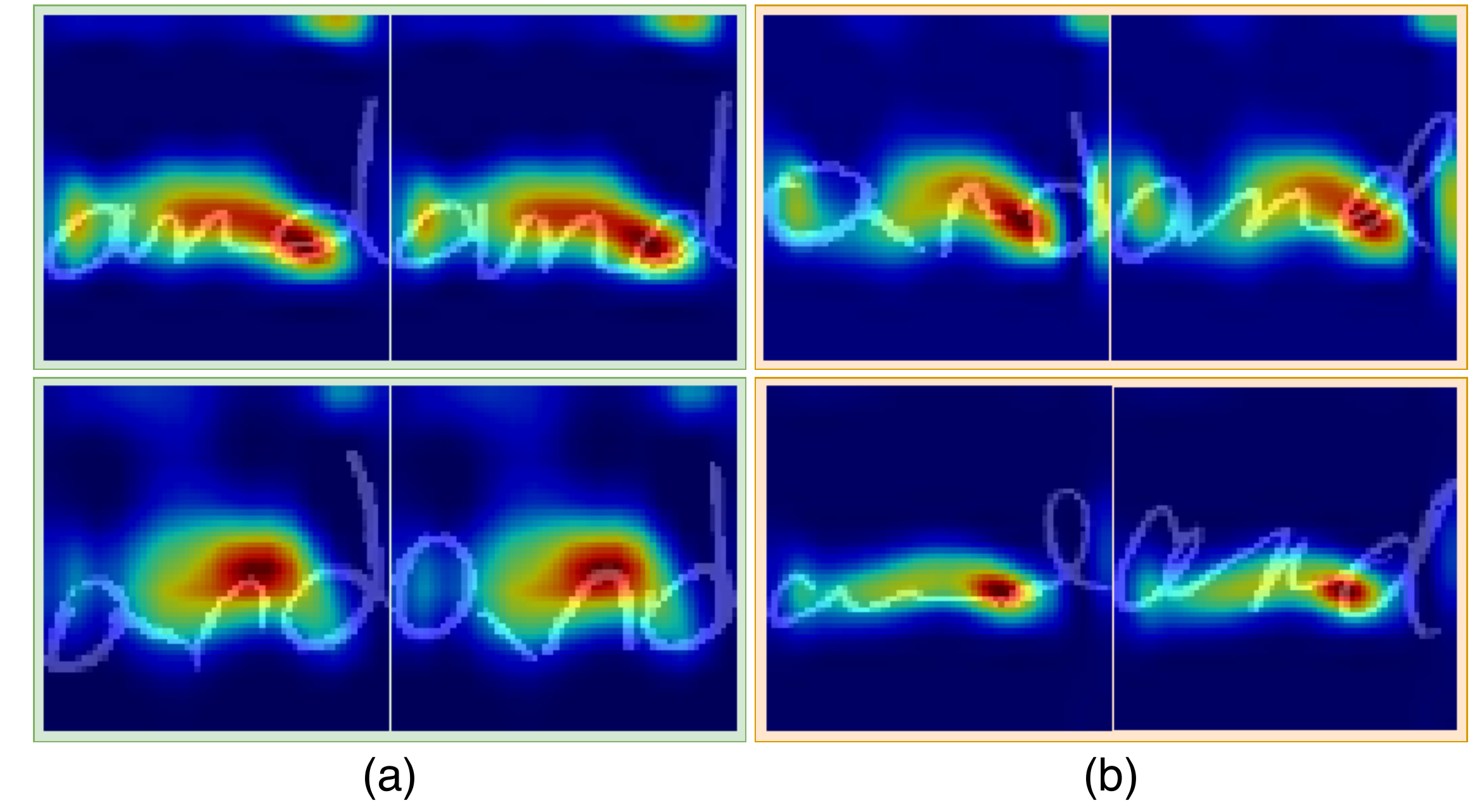}
\caption{\label{fig:sa_attentionmaps} SA Map for input images. (a) SA Maps when inputs are are from same writer, (b) SA Map when inputs are from different writers.}
\end{center}
\end{figure}
Fig. \ref{fig:sa_attentionmaps} displays energy maps from the SA layer. These maps are calculated from the output of convolutional layer connected post the CA features are extracted concatenated. Since, this SA layer was connected after the layer that outputs feature maps of shape $16 \times 16$, the shape of this SA map is $16 \times 16$. We resize these maps to $64 \times 64$ using linear interpolation and mask it on the inputs. The red and green contours show the areas that the network finds most useful for classification. Specifically, the red areas have received highest attention and the green areas have received lower attention, while the dark blue areas are mere background regions. As evident from Fig. \ref{fig:sa_attentionmaps}a top row, the energy is concentrated around the loop of ``a" and ``d" and the tent of ``n" which are very visually similar in the samples, where as in the top row of Fig. \ref{fig:sa_attentionmaps}b the energy concentrations are around dissimilar portions of ``n" and dissimilar loops of ``d". This shows that the network is able to indicate pixel regions relevant to the classification of inter and intra-class samples respectively.

We experimented with VGG16 and IRv2 and observed better results and faster convergence with VGG16. Also applying batch-normalization after each attention layer and ReLU activation after each concatenation layer improved the performance and also led to faster convergence.

\section{Conclusion}
In the domain of handwriting verification, it is required to explain the model's decision which can assist the Forensic Document Examiner (FDE). In this work, we have displayed the potential of Cross Attention and Soft Attention mechanism for this task which not only performs at par or better than the state-of-the-art methods but also provides insightful explanations on our model's decision. The CA modules can be easily set up to work with multi-modality data, i.e. to find cross-relevance in data modalities. Here, CA modules extract the intra-writer relevancies very effectively, while the SA module successfully extracts the most crucial pixel locations in the joint input. In future work we plan to apply CA to multi-modal datasets and also extend the current approach to handwritten full page datasets comprising of multiple words. We also plan to test the feasibility of SA on larger datasets for classification and related tasks.

\bibliography{main} 
\bibliographystyle{IEEEtran}

\end{document}